\documentclass[10pt,onecolumn,a4paper]{article}

\usepackage{ziplab-tech-report}
\usepackage{amsmath}
\usepackage{mathtools}
\usepackage{amssymb}
\usepackage{booktabs}
\usepackage{multirow}
\usepackage{subcaption}
\usepackage{tabularx}
\usepackage{float}
\usepackage{wrapfig}
\usepackage{algorithm}
\usepackage{algorithmic}
\usepackage{xspace}
\usepackage[pagebackref,breaklinks,colorlinks,allcolors=ziplabblue]{hyperref}
\usepackage[capitalize]{cleveref}

\usepackage{amsmath,amsfonts,bm}

\def\eqref#1{equation~\ref{#1}}

\def\1{\bm{1}}

\DeclareMathAlphabet{\mathsfit}{\encodingdefault}{\sfdefault}{m}{sl}
\SetMathAlphabet{\mathsfit}{bold}{\encodingdefault}{\sfdefault}{bx}{n}

\newcommand{\method}{History Forcing\xspace}
\techreportshorttitle{History Forcing}
\techreportproject{\url{https://ziplab.github.io/History-Forcing-Project-Page/}}
\techreportcode{\url{https://github.com/ziplab/History-Forcing}}
\techreportemail{\href{mailto:xiaoxuanhe@zju.edu.cn}{xiaoxuanhe@zju.edu.cn}\quad \href{mailto:chenfeng1271@gmail.com}{chenfeng1271@gmail.com}}
\techreportkeywords{Video World Models, On-Policy Distillation, Long Video Generation}
\techreportfootnote{* Equal contribution. $\dagger$ Corresponding author.}
\techreportlogos{%
  \techreportlogo{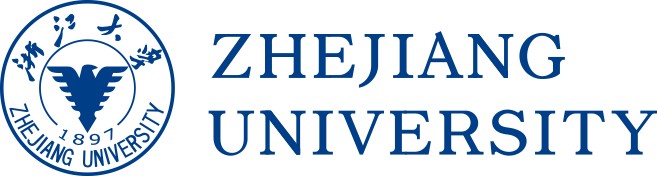}%
  \techreportlogo{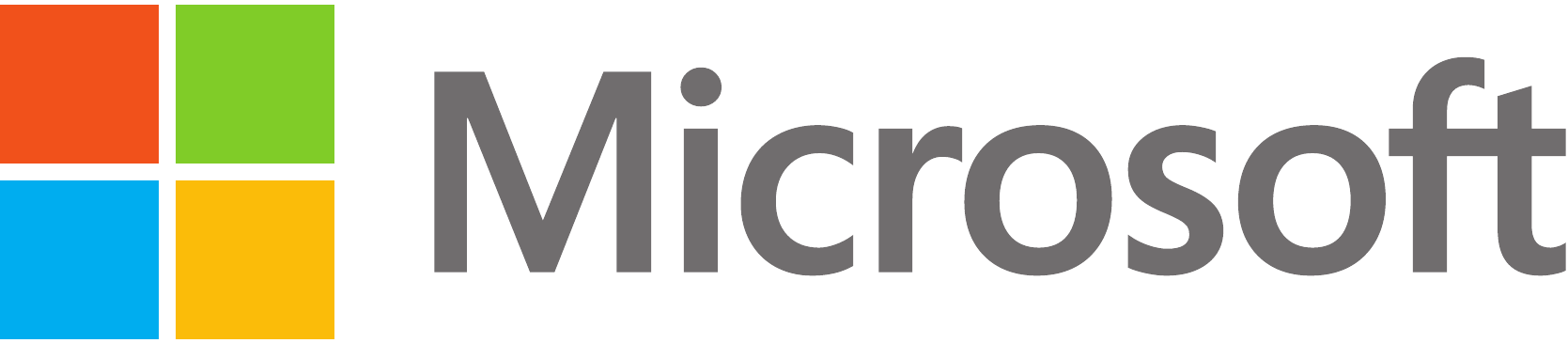}%
  \techreportlogo{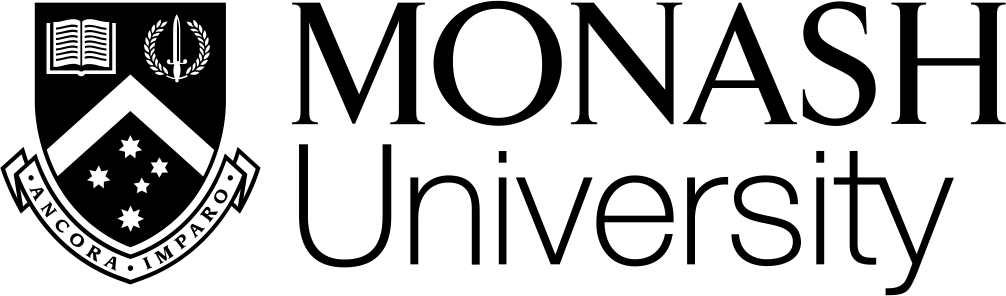}%
  \techreportlogo{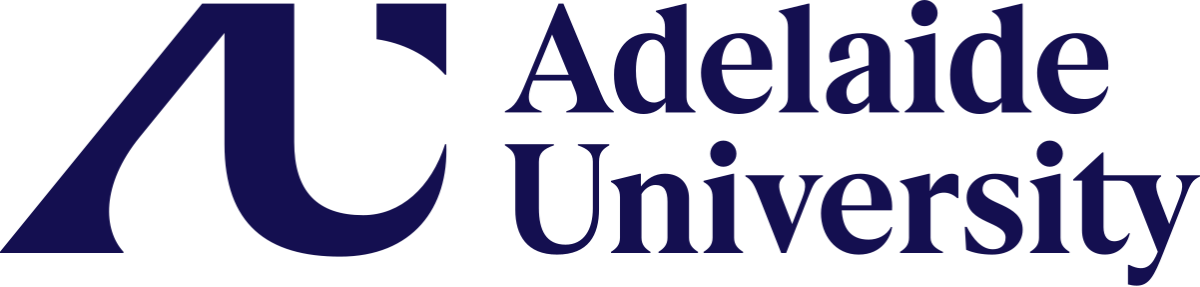}%
}

\begin{document}

\begingroup
  \ziplabmaketitle
    {History Forcing: Distilling History-Dependent Continuations for Long-Horizon Video World Models}
    {Hangyi Kuang\textsuperscript{1,*} \quad
     Chengzhi Wang\textsuperscript{1,*} \quad
     Yefei He\textsuperscript{1} \quad
     Weijie Wang\textsuperscript{1} \quad
     Yuke Zhao\textsuperscript{1,2} \quad
     Zhuokun Chen\textsuperscript{3} \quad \\
     Zisu Huang\textsuperscript{2} \quad
     Yifan Yang\textsuperscript{2} \quad
     Xiaoxuan He\textsuperscript{1,$\dagger$} \quad
     Feng Chen\textsuperscript{4,$\dagger$} \quad
     Bohan Zhuang\textsuperscript{1}}
    {\textsuperscript{1} Zhejiang University \quad
     \textsuperscript{2} Microsoft Research \quad
     \textsuperscript{3} Monash University \quad
     \textsuperscript{4} Adelaide University}
    {
      \par\noindent\begin{minipage}{\linewidth}
        \centering
        \includegraphics[width=\linewidth]{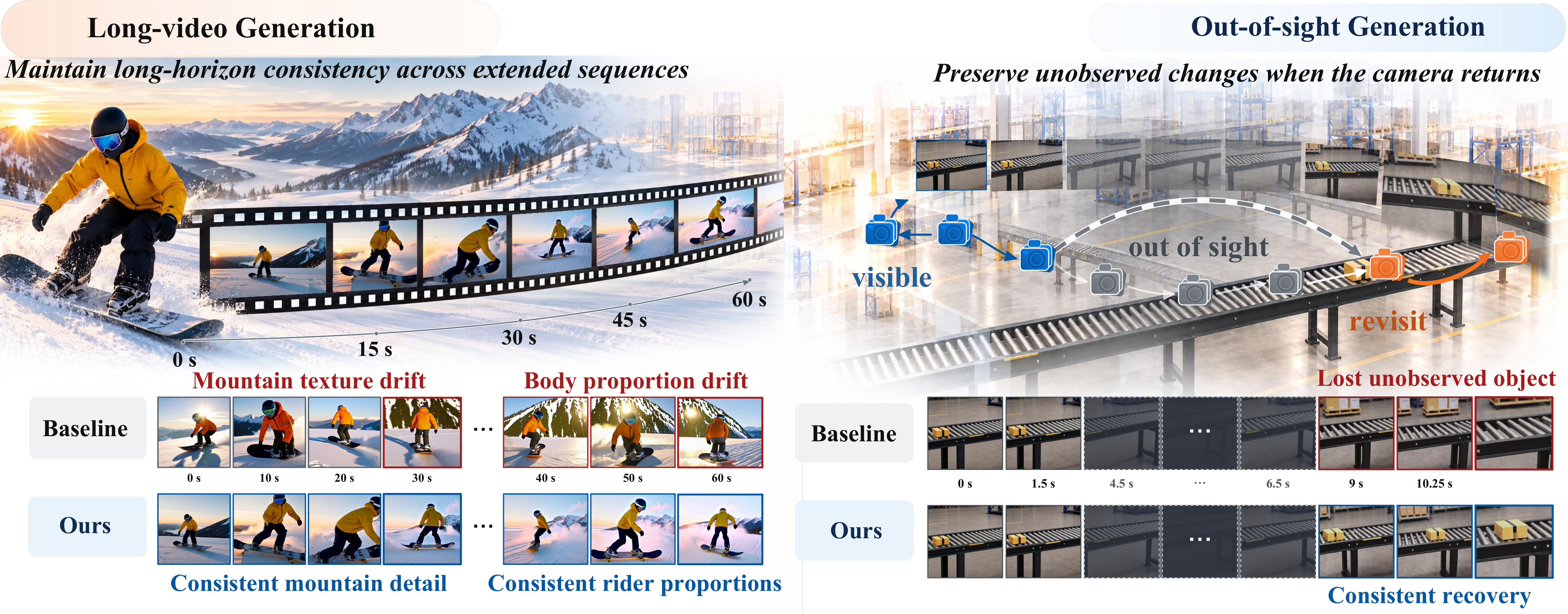}
        \captionof{figure}{%
          \method for long-horizon video world models.
          \textbf{Left: Long-video generation.} Our method maintains appearance consistency and scene continuity, while the baseline exhibits appearance drift and changes in scene layout.
          \textbf{Right: Out-of-sight world modeling.} Our method captures plausible state evolution and maintains object permanence upon revisit, while the baseline exhibits inconsistent scene changes and object disappearance.
        }
        \label{fig:teaser}
      \end{minipage}\par
      \vspace{12pt}
    }
  \renewenvironment{abstract}{}{}
  \begin{abstract}
Few-step autoregressive video diffusion enables efficient generation, yet maintaining consistency over long rollouts remains challenging, especially when subjects or scenes reappear after going out of sight. Despite progress in fidelity and rollout robustness, existing distillation methods do not fully exploit richer historical information to constrain the student's current prediction. To address this gap, we introduce \textbf{History Forcing}, a unified on-policy distillation framework that strengthens historical supervision for few-step autoregressive video generators. The framework uses the residual between teacher predictions with and without informative historical context to decouple this supervision. Such a historical context can be obtained using earlier generated chunks with less accumulated drift or a fixed-view teacher rollout that keeps the scene visible while the student looks away. The residual is then weighted and added to the score or velocity prediction used for distillation, forming a corrected target. By aligning the student with this corrected target through distillation, History Forcing incorporates long-range historical constraints into autoregressive generation. Our method achieves state-of-the-art performance among 1.3B models, with a VBench quality score of 85.07 for 15s video generation and a VBench-Long aggregate score of 79.82 for 60s generation. For out-of-sight world modeling, evaluations further demonstrate more plausible scene evolution and more consistent states when the camera returns, with Evolution Quality improving from 2.35 to 2.70 on PlayWorld and Object Consistency from 58.43 to 62.18 on MemoBench.
\end{abstract}

  \ziplabendtitle{}
\endgroup

\clearpage

\begin{figure*}[!t]
    \centering
    \includegraphics[width=\textwidth]{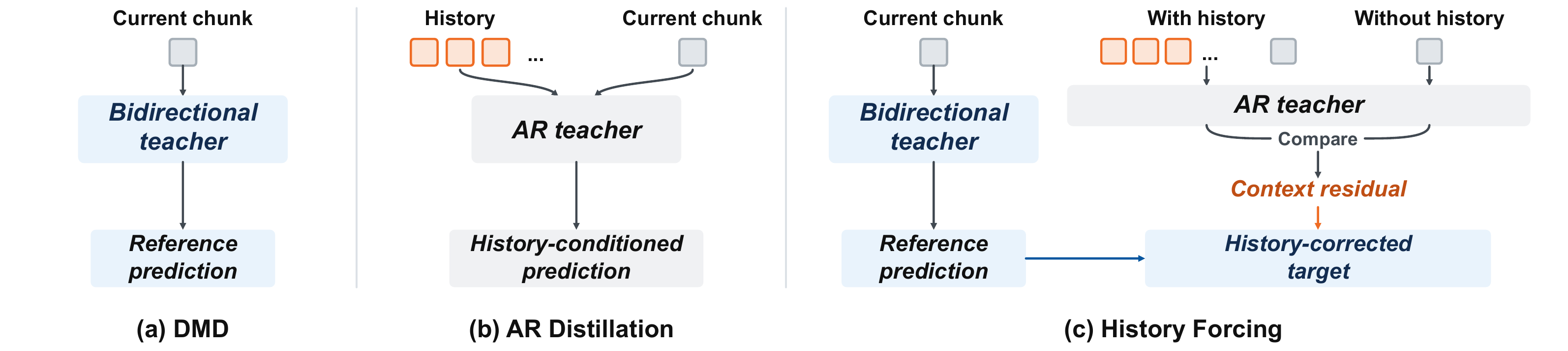}
    \caption{Comparison of distillation strategies.
    (a) DMD uses a bidirectional teacher's prediction as the reference for the current chunk.
    (b) AR distillation directly uses the history-conditioned prediction of an AR teacher as the full target.
    (c) History Forcing uses the context residual between the AR teacher's predictions with and without history to correct the reference prediction.}
    \label{fig:motivation}
\end{figure*}

\section{Introduction}
\label{sec:introduction}

Few-step autoregressive (AR) video diffusion models have enabled efficient long-horizon video generation~\citep{yin2025slow,yang2025longlive,liu2026rolling}. However, maintaining consistency over extended autoregressive rollouts remains challenging. Coherent continuation often relies on earlier observations of persistent subjects, scenes, and dynamics, yet such information may be insufficiently incorporated into later predictions, leading to gradual subject and scene drift. This reliance on long-range history becomes even more pronounced in video world models. When an entity leaves the field of view, it may reappear only after a long interval, by which point the recent frames carry no information about it~\citep{duan2026liveworld,xu2026teaching}. These challenges make effective use of long-range history critical for consistent autoregressive generation.

Existing training paradigms have progressively increased the role of historical information in autoregressive supervision. Teacher forcing conditions predictions on ground-truth prefixes~\citep{jin2025pyramidal,gao2025ca2}, whereas Self Forcing applies distribution matching distillation (DMD) to student-generated rollouts using a bidirectional diffusion teacher~\citep{huang2026self}. Subsequent methods further improve long-horizon training by extending supervision to longer generated sequences, preserving information from earlier frames, or propagating learning signals from later predictions to earlier generated representations~\citep{liu2026rolling,yang2025longlive,zhuang2026self}. More recent approaches condition teacher predictions on longer contexts, causally matched prefixes, or privileged observations, allowing the distillation target to better reflect historical dependencies relevant to autoregressive generation~\citep{chen2026context,bandyopadhyay2026context,liu2026opsd}.

Despite these advances,
existing distillation targets do not provide an explicit mechanism to strengthen historical constraints independently of the teacher’s overall generative supervision.
As illustrated in \cref{fig:motivation}, DMD with a bidirectional diffusion teacher provides strong supervision for video quality, but does not explicitly incorporate distant autoregressive history into teacher predictions. History-conditioned methods instead distill from an AR teacher with access to richer historical context~\citep{chen2026context,bandyopadhyay2026context}. However, matching its full conditional distribution couples supervision for temporal consistency and visual quality. The AR teacher's weaknesses in visual quality may therefore constrain the student~\citep{yin2025slow}. We refer to this limitation as \emph{history--quality coupling}.

To address this limitation, we introduce \textbf{History Forcing}, a unified on-policy distillation framework that isolates and strengthens historical supervision. We compare teacher predictions with and without informative historical context to compute a residual that captures the influence of history on the current prediction. This residual is then weighted and added to the score or velocity prediction used for distillation, guiding the student to generate continuations consistent with the historical observations.

We apply History Forcing to different tasks and develop corresponding history construction strategies. In long-video generation, we use earlier chunks of the student's rollout to provide subject and scene information less affected by accumulated drift. The resulting residual corrects the DMD target while retaining the bidirectional teacher's supervision for visual quality. For out-of-sight world modeling, we obtain history from a teacher rollout that keeps the camera fixed on the region to model its evolution while the region is outside the student's view. This history is used to compute the residual that guides the student's continuation upon revisit. As illustrated in \cref{fig:teaser}, History Forcing better preserves subject appearance and scene structure over long rollouts and produces more consistent scene states upon revisit, demonstrating the benefits of strengthening historical supervision.

Our contributions are summarized as follows:
\begin{itemize}
    \item We introduce History Forcing, a unified distillation framework that \emph{decouples} historical supervision from generative supervision. By comparing teacher predictions with and without informative history, we compute a context residual that is weighted and added to the original distillation target, allowing historical constraints to be strengthened independently. This formulation can be readily incorporated into DMD and on-policy self-distillation and is compatible with existing long-video training strategies.

    \item We develop history construction strategies for long-video generation and out-of-sight world modeling, using earlier generated chunks to constrain temporal drift and fixed-camera teacher rollouts to supervise scene evolution.

    \item History Forcing achieves a VBench quality score of 85.07 for 15s generation and directly extrapolates to 60s generation with a VBench-Long aggregate score of 79.82, using student training rollouts of at most 15s. With only 1.3B teachers, it outperforms the compared 1.3B models, including those distilled from 14B teachers, and approaches the 5B LongLive-2.0 model on VBench-Long. For out-of-sight world modeling, it further improves scene evolution and revisit consistency on PlayWorld and MemoBench.
\end{itemize}

\section{Related Work}
\label{sec:related_work}

\textbf{Long-horizon autoregressive video diffusion.} Existing video diffusion models provide strong visual and motion priors for short video synthesis~\citep{yang2025cogvideox,kong2024hunyuanvideo,wan2025wan,hacohen2024ltx}, while extending generation to long temporal horizons requires mitigating error accumulation and maintaining useful historical context. Training-free approaches extend pretrained models through diagonal denoising, temporal feature manipulation, and cache compression~\citep{kim2024fifo,lu2024freelong,yi2025deep}. Autoregressive approaches instead improve the rollout process through extended self-rollouts, bounded contexts, rolling denoising, or error-aware training~\citep{yang2025longlive,liu2026rolling,cui2026self,chen2024diffusion,henschel2025streamingt2v,gu2025long,yesiltepe2026infinity,li2026stable,guo2026end}. Context packing, learned memory representations, and history-aware supervision further improve access to earlier observations~\citep{zhang2025packing,mao2026packforcing,bian2026echo,yu2025videossm,zhang2025pretraining,ding2026layerrecall,zhuang2026self,xue2026ring}. History-Guided Video Diffusion combines predictions under different historical conditions during sampling~\citep{song2025history}. We instead design distillation targets that explicitly incorporate historical supervision into on-policy training.

\noindent \textbf{Few-step causal video diffusion distillation.} Few-step causal video diffusion builds on diffusion and flow distillation to reduce autoregressive denoising costs~\citep{feng2026one,gu2026anyflow,chen2026data,lu2026reward,lu2026raven,cai2026mode}. DMD and DMD2 establish distribution matching as an effective objective for few-step generation~\citep{yin2024one,yin2024improved}. CausVid uses DMD to distill a few-step AR generator from a bidirectional diffusion teacher, while Self Forcing further performs distillation on student-generated rollouts to reduce exposure bias~\citep{yin2025slow,huang2026self}. Subsequent work improves causal initialization and teacher--student alignment~\citep{zhao2026causal,zheng2026causal}: Causal Forcing uses an AR teacher for ODE initialization, while Context-Matched Distillation aligns teacher scoring with the causal context available to the student~\citep{zhu2026causal,bandyopadhyay2026context}. Context Forcing incorporates longer history into an AR teacher's conditional distribution, while OPSD-V uses real-video history to provide EMA velocity targets along student trajectories~\citep{chen2026context,liu2026opsd}. These approaches improve sampling efficiency and teacher--student alignment, but the contribution of historical context remains implicit in the target.

\noindent \textbf{History mechanisms in video world models.} Historical conditioning is also central to video world models~\citep{wu2026infinite}, where camera motion can separate a revisit from its last informative observation. Video generation provides a basis for interactive world simulation~\citep{bruce2024genie,alonso2024diffusion,valevski2025diffusion,agarwal2025cosmos,li2025hunyuan,he2025matrix,zhu2026sana}. Real-time interactive systems combine controllable autoregressive generation with long-rollout training and scene memory~\citep{team2026dreamx,gao2026advancing,zhao2026minwm}. WorldMem combines spatial and temporal cues for historical conditioning, while VMem uses geometry-indexed view retrieval to maintain cross-view consistency~\citep{xiao2026worldmem,li2025vmem,yu2025context}. Memory Forcing trains models to balance temporal and spatial memory during exploration and revisits, while WorldPlay reconstitutes historical context and aligns teacher and student memory during distillation~\citep{huang2025memory,sun2025worldplay}. LiveWorld uses historical references and fixed-view virtual monitors to preserve scene identity and continue unseen events, while ReMind trains models to recall reliable historical observations across occlusions and viewpoint changes~\citep{duan2026liveworld,xu2026teaching}. These methods improve scene consistency through memory mechanisms and history-aware training, but coherent continuation across long observation gaps remains challenging.

\section{Preliminaries}
\label{sec:preliminaries}

\paragraph{Diffusion models.}
Diffusion models learn to reverse a Gaussian perturbation process.
Given a clean latent $x_0$, its noisy state at diffusion time $t$ is
$x_t=\alpha_t x_0+\sigma_t\epsilon$, where
$\epsilon\sim\mathcal{N}(0,I)$ and $(\alpha_t,\sigma_t)$ specifies the
noise schedule. Under the linear interpolation commonly used in flow
matching, the denoising network predicts the velocity
$v_t=\epsilon-x_0$~\citep{lipman2022flow}. A predicted velocity
$\hat v_t$ gives the clean estimate
$\hat x_0=x_t-\sigma_t\hat v_t$.
Let $p_t$ denote the distribution of $x_t$. Its score is
$s(x_t,t)=\nabla_{x_t}\log p_t(x_t)$ and can be estimated from the
denoising prediction~\citep{song2020score,yin2024one}:
\begin{equation}
    s(x_t,t)
    \approx
    \frac{\alpha_t\hat x_0-x_t}{\sigma_t^2}.
    \label{eq:denoising_score}
\end{equation}
Sampling starts from Gaussian noise and iteratively updates the noisy
latent using the denoising predictions.

\paragraph{Autoregressive video diffusion.}
A pretrained 3D variational autoencoder (VAE) encodes RGB video frames
into a spatially and temporally compressed sequence of latent feature
maps, referred to as latent frames~\citep{yang2025cogvideox}.
The diffusion model operates in this latent space, and the denoised
latents are decoded into video frames using the VAE
decoder~\citep{kong2024hunyuanvideo}.
Autoregressive video diffusion models generate videos sequentially in
chunks, each comprising consecutive latent frames~\citep{gao2025ca2}.
Let $x_0^i$ denote the $i$-th clean latent chunk in a sequence of $N$
chunks. The video distribution is factorized as:
\begin{equation}
    p_{\theta}(x_0^{1:N})
    =
    \prod_{i=1}^{N}
    p_{\theta}(x_0^i\mid x_0^{<i}).
    \label{eq:ar_factorization}
\end{equation}
At inference, each chunk is generated by iteratively denoising its latent
frames, conditioned on previously generated chunks.

\paragraph{Distribution matching distillation.}
Distribution Matching Distillation (DMD) trains a few-step generator to
match a reference distribution defined by a pretrained diffusion model
~\citep{yin2024one,yin2024improved}. Let $q_{\theta,t}$ and
$p_{\mathrm{ref},t}$ denote the diffused student and reference
distributions. DMD minimizes
$\mathbb{E}_t[
D_{\mathrm{KL}}(q_{\theta,t}\Vert p_{\mathrm{ref},t})]$.
Its distribution-matching gradient is approximated by
\begin{equation}
    \nabla_{\theta}\mathcal{L}_{\mathrm{DMD}}
    \approx
    \mathbb{E}\!\left[
        w_t\alpha_t J_{\theta}^{\top}
        \bigl(s_{\mathrm{fake}}-s_{\mathrm{ref}}\bigr)
    \right],
    \label{eq:dmd_gradient}
\end{equation}
where $J_{\theta}=\partial x_0^i/\partial\theta$ and $w_t$ is the DMD
weighting factor. Here,
$s_{\mathrm{ref}}=\nabla_{x_t^i}\log p_{\mathrm{ref},t}(x_t^i)$ and
$s_{\mathrm{fake}}=\nabla_{x_t^i}\log q_{\theta,t}(x_t^i)$ denote the
reference and fake scores. The reference score is estimated by the
pretrained diffusion model, while the fake score is estimated by an online
denoiser trained on student-generated samples. Both scores are obtained
from denoising predictions using \cref{eq:denoising_score}.

\section{Method}
\label{sec:method}

\begin{figure*}[t]
    \centering
    \includegraphics[width=\linewidth]{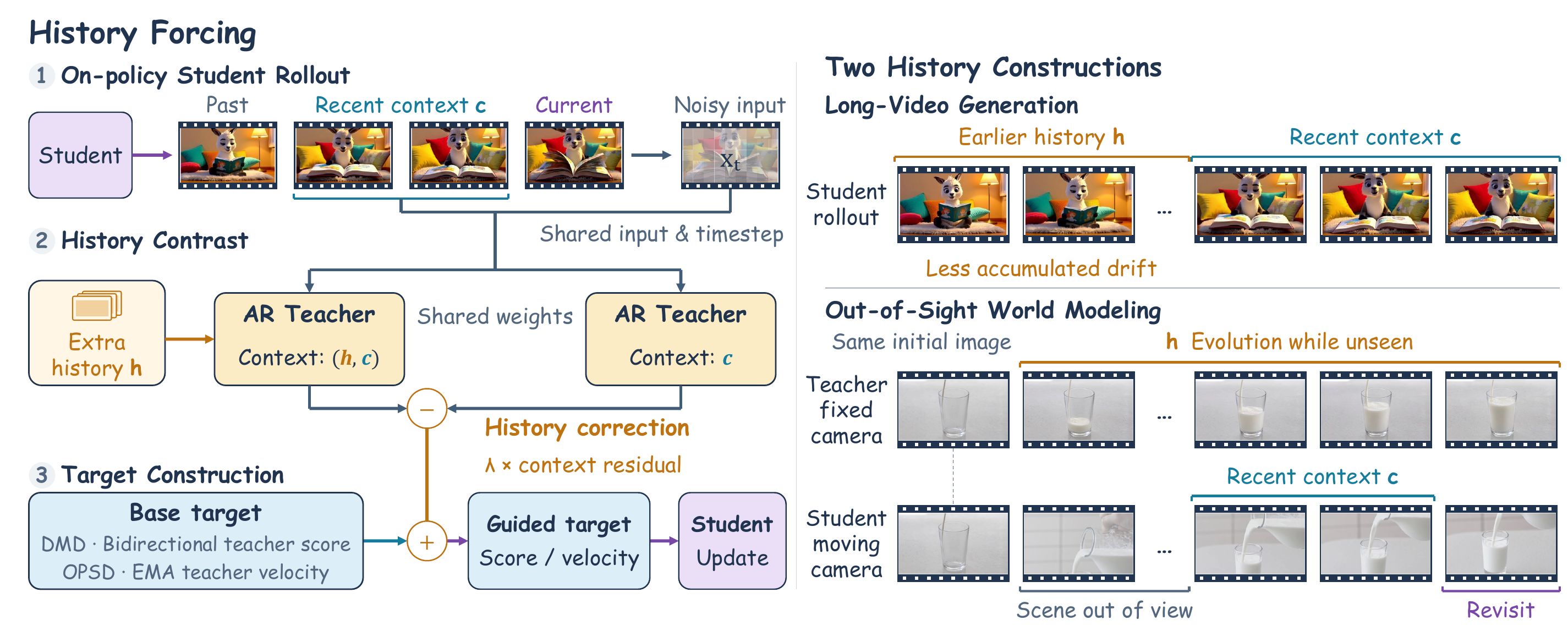}
    \caption{Overview of History Forcing. An AR teacher predicts the same noisy student chunk with and without additional history. The resulting history contrast modifies the reference score in DMD and the velocity target in on-policy self-distillation.}
    \label{fig:overview}
\end{figure*}

History Forcing derives a history-dependent correction by comparing AR teacher predictions for the same noisy student sample under two contexts that differ only in additional history. As shown in \cref{fig:overview}, this correction is applied to the reference score in DMD for long-horizon consistency and to the teacher velocity target in on-policy self-distillation for out-of-sight evolution.

\subsection{History-Reweighted Distribution}
\label{sec:history_distribution}

For the current chunk $x_0^i$, let $h^i$ denote the additional historical observations used for supervision and $c$ denote the shared condition, including the recent context and other inputs held fixed across the two AR teacher predictions. The AR diffusion teacher predicts the same noisy chunk with and without $h^i$, keeping $c$ fixed. We denote the corresponding diffused distributions by $p_{\mathrm{AR},t}^{+}=p_{\mathrm{AR},t}(\cdot\mid h^i,c)$ and $p_{\mathrm{AR},t}^{-}=p_{\mathrm{AR},t}(\cdot\mid c)$, with scores $s_{\mathrm{AR}}^{+}$ and $s_{\mathrm{AR}}^{-}$, respectively. Let $p_{\mathrm{ref},t}$ denote a reference distribution, such as that defined by a pretrained bidirectional diffusion teacher, with score $s_{\mathrm{ref}}$. Motivated by classifier-free guidance (CFG)~\citep{ho2022classifier,song2025history}, we express the history-dependent correction in score form as:
\begin{equation}
s_{\mathrm{HF}}=s_{\mathrm{ref}}+\lambda\left(s_{\mathrm{AR}}^{+}-s_{\mathrm{AR}}^{-}\right),
\label{eq:history_score}
\end{equation}
where $\lambda\geq0$ controls the strength of historical supervision. All scores are estimated at the same noisy chunk and diffusion time from the corresponding denoising predictions using \cref{eq:denoising_score}.
Using the definition of the score, \cref{eq:history_score} can be written as:
\begin{equation}
\begin{aligned}
s_{\mathrm{HF}}
&=\nabla_{x_t^i}\log p_{\mathrm{HF},t}(x_t^i)\\
&=\nabla_{x_t^i}\log\!\left[
p_{\mathrm{ref},t}(x_t^i)
\left(
\frac{p_{\mathrm{AR},t}^{+}(x_t^i)}
     {p_{\mathrm{AR},t}^{-}(x_t^i)}
\right)^{\lambda}
\right].
\end{aligned}
\label{eq:history_score_density}
\end{equation}
This gives the normalized target distribution:
\begin{equation}
p_{\mathrm{HF},t}(x_t^i)
=
\frac{1}{Z_t}\,
p_{\mathrm{ref},t}(x_t^i)
\left(
\frac{p_{\mathrm{AR},t}^{+}(x_t^i)}
     {p_{\mathrm{AR},t}^{-}(x_t^i)}
\right)^{\lambda},
\label{eq:history_target}
\end{equation}
where $Z_t$ is the partition function, which is assumed to be finite and positive. The ratio reweights the reference distribution according to how $h^i$ changes the likelihood of the noisy chunk under the AR teacher.
Treating the history-free AR distribution as the marginal over additional history, Bayes' rule gives:
\begin{equation}
\begin{gathered}
\frac{p_{\mathrm{AR},t}^{+}(x_t^i)}
     {p_{\mathrm{AR},t}^{-}(x_t^i)}
=
\frac{p_{\mathrm{AR},t}(h^i\mid x_t^i,c)}
     {p_{\mathrm{AR}}(h^i\mid c)},\\
s_{\mathrm{HF}}
=
s_{\mathrm{ref}}
+\lambda\nabla_{x_t^i}
\log p_{\mathrm{AR},t}(h^i\mid x_t^i,c).
\end{gathered}
\label{eq:history_bayes}
\end{equation}
The log-posterior gradient defines a local direction that increases the AR teacher's posterior density of the additional history, providing guidance for the reference score.

\subsection{Distillation for Long-Horizon Consistency}
\label{sec:history_dmd}

To improve long-horizon consistency, we use the history-reweighted distribution in \cref{eq:history_target} as the target for DMD. A pretrained bidirectional diffusion model defines the reference distribution, while a frozen AR teacher provides the history-dependent correction.

\begin{figure}[!t]
    \centering
    \includegraphics[width=0.7\textwidth]{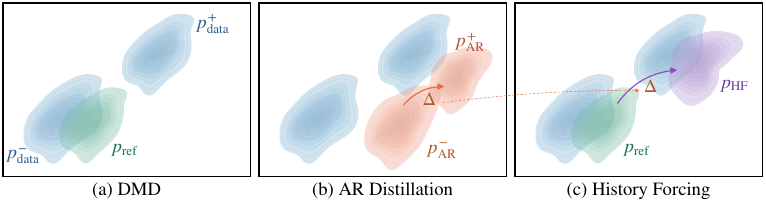}
    \caption{Comparison of distillation targets, where $+$ and $-$ denote distributions with and without additional history. \textbf{(a) DMD} uses a reference close to the data distribution without history. \textbf{(b) AR Distillation} uses the history-conditioned AR distribution, with larger modeling error. \textbf{(c) History Forcing} transfers the history-induced shift from the AR teacher to the reference, producing a target closer to the data distribution with history.}
    \label{fig:distillation_targets}
\end{figure}%

During long autoregressive rollouts, subject appearance and scene structure may gradually drift as generation errors accumulate. Earlier chunks can provide subject and scene information that is less affected by accumulated drift. To incorporate this information into distillation, we construct the additional history $h^i$ from consecutive latent frames immediately preceding the shared recent context in the same student rollout. One AR teacher context combines this additional history with the recent latent frames in chronological order, whereas the other retains only the recent frames. Sharing the recent context helps preserve autoregressive continuity and avoid an excessive gap between the two conditional distributions.

The residual between the two teacher scores provides a history-dependent correction to the bidirectional reference score. As illustrated in \cref{fig:distillation_targets}, the reference distribution closely approximates the data distribution but does not incorporate the additional history. Direct AR distillation uses the history-conditioned AR distribution as the full target, tying historical supervision to the AR teacher's generative accuracy. History Forcing instead retains the bidirectional reference score $s_{\mathrm{ref}}$ for visual-quality supervision and uses the posterior-gradient term in \cref{eq:history_bayes} to encourage consistency with the additional history, thereby decoupling historical supervision from matching the full AR distribution.

To match the history-reweighted target, the student minimizes $\mathcal{L}_{\mathrm{HF}}=\mathbb{E}_t[D_{\mathrm{KL}}(q_{\theta,t}\Vert p_{\mathrm{HF},t})]$. The history and shared condition are held fixed during each student update. Using \cref{eq:history_bayes}, minimizing the reverse KL at timestep $t$ is equivalent to maximizing
\begin{equation}
\mathcal{J}_t(\theta)
=
-D_{\mathrm{KL}}\!\left(q_{\theta,t}\Vert p_{\mathrm{ref},t}\right)
+\lambda\,\mathbb{E}_{x_t^i\sim q_{\theta,t}}
\left[\log p_{\mathrm{AR},t}(h^i\mid x_t^i,c)\right].
\label{eq:history_objective}
\end{equation}
The KL term matches the student to the bidirectional reference distribution, while the posterior term encourages consistency with the additional history.

This objective is optimized using autoregressive rollouts generated with the student's inference-time sampling procedure. Score estimates are obtained from denoising predictions on sampled chunks perturbed at random diffusion timesteps. Replacing the reference score in \cref{eq:dmd_gradient} with $s_{\mathrm{HF}}$ yields the DMD update:
\begin{equation}
\nabla_{\theta}\mathcal{L}_{\mathrm{HF}}
\approx\mathbb{E}\!\left[w_t\alpha_t J_{\theta}^{\top}(s_{\mathrm{fake}}-s_{\mathrm{HF}})\right].
\label{eq:history_gradient}
\end{equation}
The expectation is taken over student samples, diffusion timesteps, and perturbation noise. The DMD weighting and fake-score training remain unchanged.


\subsection{Refinement for Out-of-Sight Evolution}

In camera-controlled video world models, scene evolution during out-of-sight intervals poses an additional challenge to long-horizon consistency. Earlier chunks of the student's rollout provide historical observations that constrain subject appearance and scene structure upon revisit, but do not capture changes occurring outside the field of view. Historical supervision derived from these observations therefore provides limited guidance for out-of-sight evolution.

To provide supervision for out-of-sight evolution, we use a teacher model to generate a video from the same initial scene image as the student.
The student follows its prescribed camera trajectory, whereas the teacher keeps the camera at its initial pose to model how the scene evolves while outside the student's view.
As illustrated in \cref{fig:overview}, we construct $h^i$ for each supervised student chunk using latent frames from the teacher's fixed-camera rollout at timestamps preceding the start of that chunk.
To extract supervision from this history, we compare teacher predictions with and without $h^i$. The teacher's inputs share the latent frames immediately preceding the current chunk in the student's rollout to preserve continuity along the student's camera trajectory.

We refine the pretrained world model through on-policy self-distillation to improve out-of-sight evolution while preserving its visual quality.
Because a distilled few-step generator is not trained to provide accurate denoising predictions across arbitrary noise levels, directly using it as a score teacher can be unreliable.
Following OPSD-V~\citep{liu2026opsd}, we instead supervise the student's velocity predictions along its sampling trajectory using history-guided targets constructed from an exponential moving average (EMA) teacher.
This formulation requires no auxiliary fake-score model.

Let $v_{\mathrm{EMA}}^{+}$ and $v_{\mathrm{EMA}}^{-}$ denote the teacher's velocity predictions with and without $h^i$, using the same noisy student chunk, timestep, prompt, and camera controls. History Forcing constructs the corrected target as:
\begin{equation}
v_{\mathrm{HF}}
=
v_{\mathrm{EMA}}^{-}
+
\lambda\left(v_{\mathrm{EMA}}^{+}-v_{\mathrm{EMA}}^{-}\right).
\end{equation}
The student is optimized through
\begin{equation}
\mathcal{L}_{\mathrm{HF}}
=
\mathbb{E}\!\left[
\operatorname{MSE}\!\left(
v_{\theta},\operatorname{sg}[v_{\mathrm{HF}}]
\right)
\right],
\end{equation}
where $v_{\theta}$ denotes the student's velocity prediction, and $\operatorname{sg}$ denotes stop-gradient. Supervision is applied to chunks generated during revisits.

\newsavebox{\vbenchfirsttable}

\begin{table*}[!t]
\centering
\caption{Comparison on VBench for 15s video generation. Causal Forcing's released AR diffusion checkpoint~\citep{zhu2026causal}, trained on 5s videos, serves as the 5s teacher. We further fine-tune this checkpoint on 10s videos. We evaluate the released checkpoints of all competing methods. Best and second-best scores are bold and underlined.}
\label{tab:model-comparison}
\sbox{\vbenchfirsttable}{%
\begin{tabular}{@{}lcccccccc@{}}
\toprule
\textbf{Method}
& \shortstack{\textbf{Quality}\\\textbf{Score} $\uparrow$}
& \shortstack{\textbf{Subject}\\\textbf{Consistency} $\uparrow$}
& \shortstack{\textbf{Background}\\\textbf{Consistency} $\uparrow$}
& \shortstack{\textbf{Temporal}\\\textbf{Flickering} $\uparrow$}
& \shortstack{\textbf{Motion}\\\textbf{Smoothness} $\uparrow$}
& \shortstack{\textbf{Dynamic}\\\textbf{Degree} $\uparrow$}
& \shortstack{\textbf{Aesthetic}\\\textbf{Quality} $\uparrow$}
& \shortstack{\textbf{Imaging}\\\textbf{Quality} $\uparrow$} \\
\midrule
CausVid & 82.68 & 92.16 & 92.43 & 99.08 & 98.23 & 60.56 & 66.30 & 67.94 \\
Self-Forcing & 83.70 & 92.46 & 93.97 & 98.84 & 98.61 & 61.11 & 66.54 & 70.97 \\
LongLive & 84.20 & 94.15 & \underline{95.75} & 99.11 & \textbf{98.77} & 53.33 & \textbf{68.06} & 70.94 \\
Rolling Forcing & 84.20 & \textbf{95.61} & \textbf{95.80} & 98.85 & \underline{98.62} & 51.11 & 66.89 & \textbf{72.65} \\
Infinity-RoPE & 84.54 & 92.64 & 95.52 & \underline{99.25} & 98.49 & 68.89 & 66.47 & 69.65 \\
Context Forcing & 83.76 & \underline{94.20} & 95.33 & 98.77 & 98.28 & 60.28 & 66.47 & 69.32 \\
\midrule
Baseline & 84.44 & 91.84 & 93.58 & 98.99 & 98.46 & \underline{76.67} & 66.10 & 69.83 \\
\method\ (5s teacher) & \underline{84.82} & 92.58 & 93.91 & 99.05 & 98.42 & \textbf{78.06} & 66.70 & 69.64 \\
\method\ (10s teacher) & \textbf{85.07} & 93.21 & 95.03 & \textbf{99.34} & 98.56 & 70.00 & \underline{67.52} & \underline{70.98} \\
\bottomrule
\end{tabular}%
}
\global\setbox\vbenchfirsttable=\copy\vbenchfirsttable
\resizebox{\textwidth}{!}{\usebox{\vbenchfirsttable}}
\end{table*}

\begin{table*}[!t]
\centering
\caption{Comparison on VBench-Long for 60s video generation. Results for competing methods are taken from LongLive-2.0~\citep{chen2026longlive}. Aggregate Score uses VBench normalization and weights over the six metrics. Teacher Size refers to the DMD teacher. \method uses the AR teacher fine-tuned on 10s videos. Best and second-best results among 1.3B models are bold and underlined.}
\label{tab:vbench_long_60s}
\resizebox{\textwidth}{!}{%
\resizebox{\wd\vbenchfirsttable}{!}{%
\begin{tabular*}{\wd\vbenchfirsttable}{@{\extracolsep{\fill}}l*{9}{@{}c}@{}}
\toprule
\textbf{Method}
& \shortstack{\textbf{Student}\\\textbf{Size}}
& \shortstack{\textbf{Teacher}\\\textbf{Size}}
& \shortstack{\textbf{Aggregate}\\\textbf{Score} $\uparrow$}
& \shortstack{\textbf{Subject}\\\textbf{Consistency} $\uparrow$}
& \shortstack{\textbf{Background}\\\textbf{Consistency} $\uparrow$}
& \shortstack{\textbf{Motion}\\\textbf{Smoothness} $\uparrow$}
& \shortstack{\textbf{Dynamic}\\\textbf{Degree} $\uparrow$}
& \shortstack{\textbf{Aesthetic}\\\textbf{Quality} $\uparrow$}
& \shortstack{\textbf{Imaging}\\\textbf{Quality} $\uparrow$} \\
\midrule
\multicolumn{10}{@{}l@{}}{\textit{Other model sizes}} \\
NOVA & 0.6B & --- & 64.22 & 77.50 & 88.06 & 98.94 & 12.00 & 47.53 & 44.97 \\
MAGI-1 & 4.5B & --- & 71.33 & 79.46 & 87.76 & 99.26 & 56.00 & 52.10 & 54.54 \\
LongLive-2.0 & 5B & 5B & 79.90 & 97.48 & 97.00 & 98.86 & 60.62 & 53.68 & 65.51 \\
\midrule
\multicolumn{10}{@{}l@{}}{\textit{1.3B models}} \\
Causal Forcing & 1.3B & 14B & 76.43 & 93.52 & 94.12 & 95.74 & \textbf{72.32} & 51.24 & 62.30 \\
SkyReels-V2 & 1.3B & --- & 74.80 & 84.99 & 89.95 & \textbf{98.67} & 44.00 & 57.64 & 66.67 \\
Self-Forcing & 1.3B & 14B & 77.73 & 95.84 & 95.27 & 98.20 & 51.72 & 56.05 & 62.22 \\
CausVid & 1.3B & 1.3B & 76.85 & 86.75 & 89.85 & 98.47 & \underline{52.00} & \underline{62.88} & 67.47 \\
Rolling Forcing & 1.3B & 14B & \underline{79.22} & 94.09 & 94.47 & \underline{98.65} & 36.00 & \textbf{63.50} & \textbf{72.42} \\
LongLive & 1.3B & 14B & 79.12 & \underline{97.13} & \underline{95.89} & 98.61 & 44.56 & 58.17 & 67.56 \\
\midrule
\method & 1.3B & 1.3B & \textbf{79.82} & \textbf{97.32} & \textbf{96.19} & 98.46 & 49.06 & 58.06 & \underline{69.17} \\
\bottomrule
\end{tabular*}%
}%
}
\end{table*}

\section{Experiments}
\label{sec:experiments}

\subsection{Experimental Setup}
\label{sec:experimental-setup}

\paragraph{Long-video generation.}
We initialize the bidirectional DMD teacher and the trainable fake-score model from pretrained Wan2.1-T2V-1.3B~\citep{wan2025wan} weights.
The AR student is initialized from the released causal ODE checkpoint of Causal Forcing~\citep{zhu2026causal}.
The AR teacher is initialized from Causal Forcing's released multi-step AR diffusion checkpoint, which was trained on 5s video clips.
To assess the gains in student performance from improving the AR teacher's ability to model long-range dependencies, we further fine-tune this teacher on 10s clips from UltraVideo-Long~\citep{xue2025ultravideo}.
We train the student for 300 iterations on filtered and expanded VidProM~\citep{wang2024vidprom} prompts.

Student training follows the long-rollout training strategy of Self-Forcing++~\citep{cui2026self}.
Rollout durations are uniformly sampled between 5s and 15s, with DMD supervision applied to the final 5s window.
Both AR teacher inputs share the latest 6 student-generated latent frames.
The history-conditioned input additionally includes the 12 latent frames immediately preceding this shared context.
For efficiency, history supervision is restricted to 1 randomly sampled eligible chunk.
Unless otherwise stated, historical supervision is applied with $\lambda=1$ at diffusion timesteps $t<750$.
Our baseline follows the Self-Forcing++ training strategy, using the same student initialization and training setup as History Forcing but without historical supervision.
We evaluate 15s generation using VBench~\citep{huang2024vbench}.
Following LongLive-2.0~\citep{chen2026longlive}, we use VBench-Long~\citep{huang2025vbench++} to evaluate 60s videos generated from MovieGenBench prompts~\citep{polyak2024movie}.
See \cref{sec:appendix-long-video} for details.

\paragraph{Out-of-sight world modeling.}
We initialize the world model from the released DreamX-World-5B~\citep{team2026dreamx} checkpoint, which also serves as the baseline.
The training set contains 200 initial scene images from Pexels~\citep{pexels}, each used with 4 camera trajectories that leave and return to the initial viewpoint, yielding 800 training samples.
Our adapted OPSD-V~\citep{liu2026opsd} baseline uses fixed-view teacher rollouts in place of real-video teacher history and corresponds to the $\lambda=1$ case of History Forcing.
We fine-tune the model for 100 steps using LoRA, with supervision at all 4 denoising steps of the revisit chunks.
Unless otherwise stated, History Forcing uses $\lambda=3$.
We evaluate the models on the PlayWorld Out-of-sight subset (43 cases)~\citep{ding2026playworld} and MemoBench Real subset (164 clips)~\citep{chen2026memobench} using preset camera trajectories.
See \cref{sec:appendix-world-model} for details.

\subsection{Quantitative Comparison}
\label{sec:main-results}

For 15s generation, we compare History Forcing with CausVid~\citep{yin2025slow}, Self-Forcing~\citep{huang2026self}, LongLive~\citep{yang2025longlive}, Rolling Forcing~\citep{liu2026rolling}, Infinity-RoPE~\citep{yesiltepe2026infinity}, and Context Forcing~\citep{chen2026context}.
For 60s generation, we compare against the methods reported in LongLive-2.0~\citep{chen2026longlive}, covering student model sizes from 0.6B to 5B.
For out-of-sight world modeling, the baselines are DreamX-World-5B~\citep{team2026dreamx} and its adapted OPSD-V variant~\citep{liu2026opsd}.

\begin{table}[!t]
\noindent\begin{minipage}{\textwidth}
\centering
\begin{minipage}[t]{0.46\textwidth}
\vspace{0pt}
\centering
\caption{World-model comparison on the PlayWorld Out-of-sight subset and MemoBench Real subset. OPSD-V is adapted using fixed-view rollout history and corresponds to $\lambda=1$. PlayWorld metrics use a 1--5 rating scale. Bold indicates the best result.}
\label{tab:playworld-memobench}
\begingroup
\small
\setlength{\tabcolsep}{5pt}
\renewcommand{\arraystretch}{1.06}
\resizebox{\columnwidth}{!}{%
\begin{tabular}[c]{@{\extracolsep{\fill}}lccc@{}}
\toprule
\textbf{Metric} & \textbf{Baseline} & \textbf{OPSD-V} & \textbf{\method}\\
\midrule
\multicolumn{4}{@{}l@{}}{\textit{PlayWorld Out-of-sight}}\\
Raw Mean $\uparrow$          & 2.38 & 2.44 & \textbf{2.69}\\
Gated Mean $\uparrow$        & 2.29 & 2.37 & \textbf{2.49}\\
Evolution Quality $\uparrow$ & 2.35 & 2.42 & \textbf{2.70}\\
Subject Identity $\uparrow$  & 2.70 & 2.72 & \textbf{2.98}\\
\midrule
\multicolumn{4}{@{}l@{}}{\textit{MemoBench Real}}\\
VisQual $\uparrow$           & 50.41 & 50.62 & \textbf{50.73}\\
MotSmooth $\uparrow$         & 64.03 & 65.71 & \textbf{66.56}\\
ObjConsist $\uparrow$        & 58.43 & 59.73 & \textbf{62.18}\\
3DConsist $\uparrow$         & 84.35 & 84.51 & \textbf{86.22}\\
ORS $\uparrow$               & 62.57 & 62.83 & \textbf{62.89}\\
PSNR (dB) $\uparrow$         & 14.2971 & 14.3888 & \textbf{14.4671}\\
SSIM $\uparrow$              & 0.5363  & \textbf{0.5487} & 0.5463\\
LPIPS $\downarrow$           & 0.5703  & 0.5674 & \textbf{0.5558}\\
ImgReward $\uparrow$         & 52.27 & 52.91 & \textbf{53.43}\\
\bottomrule
\end{tabular}%
}
\endgroup
\end{minipage}\hfill
\begin{minipage}[t]{\dimexpr\textwidth-\columnsep-0.46\textwidth\relax}
\vspace{0pt}
\centering
\caption{Ablation study on historical supervision strength.}
\label{tab:history-supervision-strength}
\begingroup
\small
\setlength{\tabcolsep}{2pt}
\renewcommand{\arraystretch}{1.08}
\resizebox{\columnwidth}{!}{%
\begin{tabular}{@{}lcccc@{}}
\toprule
\textbf{Setting}
& \shortstack{\textbf{Quality}\\\textbf{Score} $\uparrow$}
& \shortstack{\textbf{Subject}\\\textbf{Consistency} $\uparrow$}
& \shortstack{\textbf{Background}\\\textbf{Consistency} $\uparrow$}
& \shortstack{\textbf{Dynamic}\\\textbf{Degree} $\uparrow$}\\
\midrule
\multicolumn{5}{@{}l@{}}{\emph{Timestep threshold}}\\
$\tau=1000$ & 84.62 & 92.35 & 93.45 & 80.28\\
$\tau=750$ & 84.82 & 92.58 & 93.91 & 78.06\\
$\tau=500$ & 85.08 & 92.52 & 94.01 & 80.00\\
\midrule
\multicolumn{5}{@{}l@{}}{\emph{History weight}}\\
$\lambda=0$ (Baseline) & 84.44 & 91.84 & 93.58 & 76.67\\
$\lambda=0.5$ & 84.68 & 91.17 & 93.16 & 89.72\\
$\lambda=1$ & 84.82 & 92.58 & 93.91 & 78.06\\
$\lambda=2$ & 84.50 & 93.29 & 93.97 & 73.61\\
\bottomrule
\end{tabular}%
}
\endgroup
\par\vspace{\floatsep}
\centering
\caption{Ablation study on history construction.}
\label{tab:history-construction}
\begingroup
\setlength{\tabcolsep}{2pt}
\renewcommand{\arraystretch}{1.08}
\resizebox{\columnwidth}{!}{%
\begin{tabular}{@{}llcccc@{}}
\toprule
\shortstack{\textbf{History}\\$(+)$}
& \shortstack{\textbf{History}\\$(-)$}
& \shortstack{\textbf{Quality}\\\textbf{Score} $\uparrow$}
& \shortstack{\textbf{Subject}\\\textbf{Consistency} $\uparrow$}
& \shortstack{\textbf{Background}\\\textbf{Consistency} $\uparrow$}
& \shortstack{\textbf{Dynamic}\\\textbf{Degree} $\uparrow$}\\
\midrule
18 recent & 6 recent & 84.82 & 92.58 & 93.91 & 78.06\\
18 recent & None & 81.91 & 90.68 & 93.50 & 65.28\\
3 sink, 15 recent & 6 recent & 83.81 & 89.78 & 92.00 & 95.83\\
\bottomrule
\end{tabular}%
}
\endgroup
\end{minipage}
\end{minipage}
\end{table}

\noindent\textbf{15-second generation.}
As shown in \cref{tab:model-comparison}, History Forcing achieves the highest quality score among the compared methods and improves temporal consistency over the DMD baseline.
Adding historical supervision from the AR teacher raises the quality score from 84.44 to 84.82 and improves subject consistency, background consistency, and dynamic degree.
Fine-tuning this teacher on 10s videos further raises the quality score to 85.07, with additional gains in both consistency measures.
These additional gains suggest that improving the AR teacher's ability to capture long-range dependencies strengthens historical supervision during distillation.

\noindent\textbf{60-second generation.}
In \cref{tab:vbench_long_60s}, History Forcing achieves the highest aggregate score of 79.82 among the compared 1.3B students on VBench-Long, exceeding Rolling Forcing by 0.60 points.
It also achieves the best subject and background consistency in this group, indicating stronger preservation of subject identity and scene appearance over minute-long rollouts.
These results are obtained with student training rollouts of at most 15s, compared with the 60s rollouts used in LongLive's streaming long tuning.
Using a 1.3B DMD teacher, our method achieves a higher aggregate score than Rolling Forcing and LongLive, which use 14B DMD teachers.
Its aggregate score also approaches that of the 5B LongLive-2.0 model (79.90).

\noindent\textbf{Out-of-sight evaluation.}
\cref{tab:playworld-memobench} shows that History Forcing improves camera-controlled revisits on both PlayWorld Out-of-sight and MemoBench Real.
On PlayWorld, it achieves the best results across all 4 metrics, with Evolution Quality and Subject Identity exceeding OPSD-V by 0.28 and 0.26 points, respectively.
These gains indicate more plausible scene progression during unobserved intervals while preserving subject identity upon revisit.
On MemoBench Real, History Forcing improves all 9 reported metrics over the baseline and outperforms OPSD-V on 8 of them.
The improvements in visual quality are accompanied by better object preservation and higher fidelity to the reference videos, suggesting that historical supervision benefits both the realism of generated frames and the consistency of scene content across viewpoint changes.

\begin{table*}[!t]
\centering
\caption{Ablation study on historical supervision strength for video world modeling.}
\label{tab:world-model-supervision-strength}
\begingroup
\small
\setlength{\tabcolsep}{2pt}
\renewcommand{\arraystretch}{1.08}
\resizebox{\textwidth}{!}{%
\begin{tabular}[c]{@{\extracolsep{\fill}}l*{13}{c}@{}}
\toprule
\textbf{Setting}
& \multicolumn{4}{c}{\textbf{PlayWorld Out-of-sight}}
& \multicolumn{9}{c}{\textbf{MemoBench Real}}\\
\cmidrule(lr){2-5}\cmidrule(lr){6-14}
& \shortstack{Raw\\Mean $\uparrow$}
& \shortstack{Gated\\Mean $\uparrow$}
& \shortstack{Evo.\\Qual. $\uparrow$}
& \shortstack{Subject\\Identity $\uparrow$}
& \shortstack{VisQual\\$\uparrow$}
& \shortstack{Mot.\\Smooth $\uparrow$}
& \shortstack{Obj.\\Consist $\uparrow$}
& \shortstack{3D\\Consist $\uparrow$}
& \shortstack{ORS\\$\uparrow$}
& \shortstack{PSNR\\(dB) $\uparrow$}
& \shortstack{SSIM\\$\uparrow$}
& \shortstack{LPIPS\\$\downarrow$}
& \shortstack{Img.\\Reward $\uparrow$}\\
\midrule
\multicolumn{14}{@{}l@{}}{\emph{Supervised denoising steps}}\\
All 4 steps
& \textbf{2.60} & \textbf{2.51} & \textbf{2.60} & \textbf{2.86}
& 50.64 & \textbf{66.38} & \textbf{61.82} & \textbf{85.28}
& \textbf{61.66} & \textbf{14.4463} & \textbf{0.5480}
& \textbf{0.5595} & \textbf{53.36}\\
Last 3 steps
& 2.28 & 2.07 & 2.23 & 2.51
& \textbf{50.69} & 65.49 & 59.82 & 84.97
& 61.26 & 14.3745 & 0.5477 & 0.5653 & 52.88\\
Last 2 steps
& 2.45 & 2.36 & 2.44 & 2.65
& 50.61 & 64.86 & 59.26 & 84.24
& 61.19 & 14.3600 & 0.5476 & 0.5667 & 53.02\\
\midrule
\multicolumn{14}{@{}l@{}}{\emph{History weight}}\\
$\lambda=1$
& 2.44 & 2.37 & 2.42 & 2.72
& 50.62 & 65.71 & 59.73 & 84.51
& 62.83 & 14.3888 & \textbf{0.5487} & 0.5674 & 52.91\\
$\lambda=2$
& 2.60 & \textbf{2.51} & 2.60 & 2.86
& 50.64 & 66.38 & 61.82 & 85.28
& 61.66 & 14.4463 & 0.5480 & 0.5595 & 53.36\\
$\lambda=3$
& \textbf{2.69} & 2.49 & \textbf{2.70} & \textbf{2.98}
& \textbf{50.73} & \textbf{66.56} & 62.18 & 86.22
& \textbf{62.89} & 14.4671 & 0.5463
& 0.5558 & 53.43\\
$\lambda=5$
& 2.63 & 2.41 & 2.65 & 2.95
& 50.62 & 66.43 & \textbf{62.78} & \textbf{86.96}
& 62.26 & \textbf{14.4680} & 0.5407
& \textbf{0.5534} & \textbf{53.73}\\
\bottomrule
\end{tabular}%
}
\endgroup
\par\vspace{\floatsep}
\begingroup
\captionsetup{font=footnotesize}
\noindent\begin{minipage}[t]{\dimexpr\textwidth-\columnsep-0.43\textwidth\relax}
\vspace{0pt}
\centering
\caption{Ablation study on gradient normalization.}
\label{tab:normalization}
\begingroup
\small
\setlength{\tabcolsep}{2pt}
\renewcommand{\arraystretch}{1.08}
\resizebox{\linewidth}{!}{%
\begin{tabular}{@{}lcccc@{}}
\toprule
\textbf{Setting}
& \shortstack{\textbf{Quality}\\\textbf{Score} $\uparrow$}
& \shortstack{\textbf{Subject}\\\textbf{Consistency} $\uparrow$}
& \shortstack{\textbf{Background}\\\textbf{Consistency} $\uparrow$}
& \shortstack{\textbf{Dynamic}\\\textbf{Degree} $\uparrow$}\\
\midrule
Without joint norm. & 84.09 & 93.78 & 94.46 & 59.72\\
Joint norm. & 84.82 & 92.58 & 93.91 & 78.06\\
\bottomrule
\end{tabular}%
}
\endgroup
\end{minipage}\hfill
\begin{minipage}[t]{0.43\textwidth}
\vspace{0pt}
\centering
\caption{Training efficiency for long-video generation.}
\label{tab:training-efficiency}
\begingroup
\small
\setlength{\tabcolsep}{2pt}
\renewcommand{\arraystretch}{1.0}
\resizebox{\linewidth}{!}{%
\begin{tabular}{@{}lcc@{}}
\toprule
\textbf{Method}
& \textbf{Training Time} $\downarrow$
& \textbf{Peak Allocated Memory} $\downarrow$\\
& (s/step) & (GiB/GPU)\\
\midrule
DMD & 171.30 & 33.10\\
Ours & 175.26~(+2.31\%) & 33.76~(+1.99\%)\\
\bottomrule
\end{tabular}%
}
\endgroup
\end{minipage}
\par
\endgroup
\end{table*}%

\subsection{Ablation Studies}
\label{sec:ablation-studies}

We ablate the timestep threshold, history weight, teacher history construction, and the normalization scheme for long-video generation, as well as the number of supervised denoising steps and history weight for world-model self-distillation.
The long-video ablations use Causal Forcing's released multistep AR diffusion checkpoint for historical supervision and report VBench results for 15s generation.

\begin{figure*}[!t]
    \centering
    \includegraphics[width=0.95\textwidth]{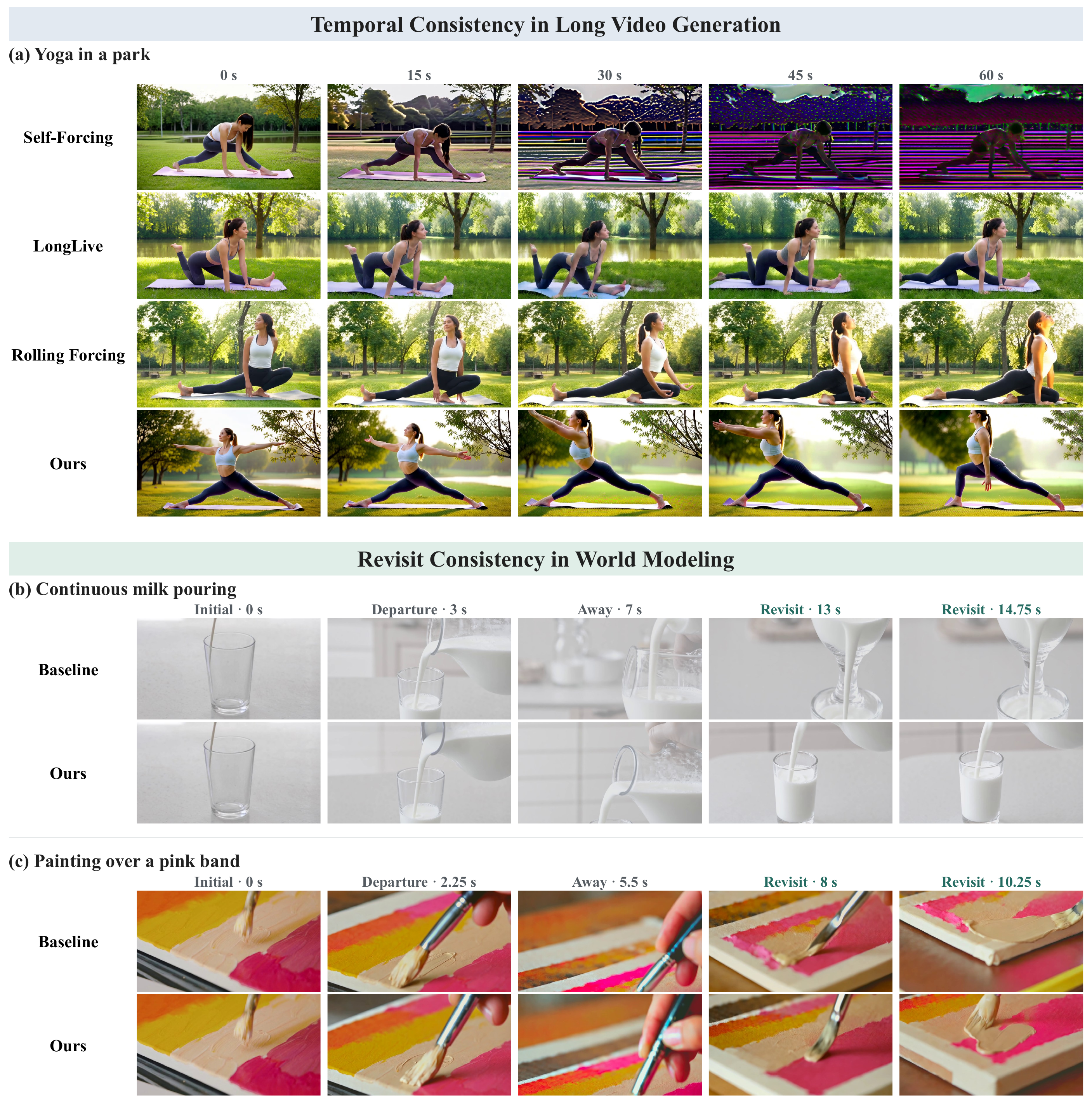}
    \caption{Qualitative comparisons of consistency. \textbf{(a)} Temporal consistency in 60s autoregressive video generation with Self-Forcing, LongLive, Rolling Forcing, and \method. \textbf{(b,c)} Revisit consistency in video world models, with each pair showing the baseline and \method.}
    \label{fig:qualitative-comparison}
\end{figure*}%

\noindent\textbf{Historical supervision strength.}
The timestep threshold $\tau$ restricts historical supervision to $t<\tau$, while $\lambda$ controls its contribution to the DMD target.
\cref{tab:history-supervision-strength} varies each factor separately, keeping the other at its default value of $\tau=750$ or $\lambda=1$.
Reducing $\tau$ from 1000 to 500 raises the quality score from 84.62 to 85.08, whereas $\tau=750$ achieves the highest subject consistency (92.58).
These results favor concentrating historical supervision at lower noise levels.
In the weight ablation, increasing $\lambda$ from 0.5 to 2 improves subject and background consistency but reduces dynamic degree.
The highest quality score is achieved at $\lambda=1$, indicating that a moderate history weight balances historical constraints with the bidirectional teacher's supervision for visual quality.

In world-model self-distillation, \cref{tab:world-model-supervision-strength} studies the number of supervised denoising steps and the history weight.
With $\lambda=2$ fixed, supervising all 4 steps yields better overall results across both benchmarks than supervising only the last 3 or 2 steps, supporting historical supervision throughout denoising.
At $\lambda=1$, the distillation target is the teacher's velocity prediction conditioned on additional history.
For out-of-sight revisits, recent student observations provide little information about the revisited region, making this history essential for constraining its identity and evolving state.
Using $\lambda>1$ amplifies the history-induced prediction change, placing greater emphasis on these constraints during self-distillation.
The history-weight ablation favors $\lambda=3$ across both benchmarks, indicating that stronger historical supervision benefits scene consistency and evolution across viewpoint changes.

\noindent\textbf{History construction.}
\cref{tab:history-construction} studies the roles of shared recent context and initial sink frames.
In the default setting, both AR teacher inputs share the latest 6 student-generated latent frames, while the $+$ input additionally includes the 12 latent frames immediately preceding this shared context.
Removing the 6 recent frames from the $-$ input reduces the quality score from 84.82 to 81.91 and lowers both consistency measures.
Sharing recent context keeps the immediate history fixed, allowing the residual to capture the contribution of earlier observations.
At the same history budget, replacing the oldest 3 frames in the $+$ input with initial sink frames also lowers the quality score to 83.81 and reduces subject and background consistency.
These results favor retaining consecutive earlier chunks for historical supervision, together with a shared recent context that preserves autoregressive continuity.

\noindent\textbf{Gradient normalization.}
\cref{tab:normalization} examines whether the history term is jointly normalized with the DMD gradient.
Standard DMD divides its distillation gradient by the mean absolute difference between the student-generated sample and the bidirectional teacher's denoising prediction.
Without joint normalization, the history term is added after this division.
Joint normalization instead computes the normalization factor using the denoising target that includes the history term, then divides the combined DMD and history update by this factor.
This increases the quality score from 84.09 to 84.82 and dynamic degree from 59.72 to 78.06, while the alternative yields higher subject and background consistency.
These results favor joint normalization for balancing temporal consistency with motion dynamics.

\noindent\textbf{Training efficiency.}
\cref{tab:training-efficiency} reports the training overhead relative to DMD.
By restricting historical supervision to one sampled chunk, we keep the increases in training time per step and peak allocated memory to only 2.31\% and 1.99\%, respectively.
History Forcing thus achieves its quality and consistency gains with modest additional training cost.

\subsection{Qualitative Comparison}
\label{sec:qualitative-comparison}

\cref{fig:qualitative-comparison} presents qualitative comparisons of temporal consistency in long-video generation and revisit consistency in out-of-sight world modeling.
In the 60s yoga example (a), Self-Forcing develops pronounced color artifacts as generation progresses.
LongLive exhibits limb duplication, while Rolling Forcing shows leg deformations during pose transitions.
\method maintains more consistent body structure and clothing appearance while preserving the park setting throughout the rollout.
For camera-controlled revisits (b,c), the baseline alters the drinking glass geometry and the colored-band layout.
In contrast, \method retains the original glass shape and preserves the surrounding pattern as paint covers the pink region.
Additional qualitative examples are provided in \cref{sec:appendix-qualitative-analysis}.

\section{Conclusion}
\label{sec:conclusion}

We presented History Forcing, which decouples historical supervision for few-step autoregressive video generators by comparing teacher predictions with and without informative history.
The method augments DMD with long-range historical constraints while retaining the bidirectional teacher's supervision for visual quality.
Fixed-view teacher history further guides out-of-sight evolution through the OPSD velocity target.
Experiments show that History Forcing reduces appearance and layout drift in long videos and better preserves object-state changes across out-of-sight intervals, producing more consistent states when the camera returns.

\paragraph{Limitations and Future Work.}
Our experiments are limited to 1.3B teachers and students for long-video generation and a 5B backbone for world modeling.
Across both settings, historical supervision remains constrained by the underlying models' ability to capture temporal dynamics.
In world modeling, for example, limited scene evolution in fixed-view teacher rollouts can weaken supervision for out-of-sight state changes.
Future work will explore larger teachers and students, longer-context training, and improvements to the world model's ability to simulate physically plausible dynamics.

{%
  \small
  \bibliographystyle{ieeenat_fullname}
  \bibliography{main}
}

\clearpage
\appendix
\crefname{appendix}{Appendix}{Appendices}
\section*{Appendix}
\noindent In the Appendix, we provide the following:
\begin{itemize}
    \item additional training, inference, and evaluation details in \cref{app:implementation-details}.
    \item additional qualitative comparisons and examples of long-video generation and out-of-sight world modeling in \cref{app:qualitative-analysis}.
\end{itemize}

\section{Additional Implementation Details}
\label{sec:appendix-implementation-details}
\label[appendix]{app:implementation-details}

Both applications use AdamW with betas $(0,0.999)$, weight decay 0.01, and a maximum gradient norm of 10.
Training uses 8 NVIDIA A100 GPUs with 1 sample per GPU and random seed 42.
Student sampling uses 4 SDE steps with unshifted timesteps $(1000,750,500,250)$ and a timestep shift of 5.
Each autoregressive chunk contains 3 latent frames, and videos are generated at 16 FPS.
All context lengths below are measured in latent frames.

\subsection{Long-Video Generation}
\label{sec:appendix-long-video}

\paragraph{Training.}
The student learning rate is $2\times10^{-6}$, and the fake-score model learning rate is $4\times10^{-7}$.
The bidirectional and AR teachers remain frozen during student distillation.
The bidirectional teacher uses a classifier-free guidance scale of 3.0.
The fake-score model is updated at every iteration and the student once every 5 iterations.
Gradient accumulation over 8 steps gives an effective batch size of 64.
The student EMA uses a decay of 0.99 and is initialized at iteration 200.
The chunk for historical supervision is sampled from the DMD supervision window.
AR teacher fine-tuning uses teacher forcing with a learning rate of $2\times10^{-6}$ and an effective batch size of 8.
For the 10s AR teacher, we use the fine-tuning checkpoint at iteration 2500.
The AR teachers are configured with attention windows of 21 and 42 latent frames for the 5s and 10s variants, respectively.
The two predictions use separate KV caches, with matched temporal positions for the current chunk.
The threshold is applied to the shifted DMD scoring timestep.
Joint normalization is used by default.

\paragraph{Inference.}
Inference uses student EMA weights at $832\times480$ resolution.
All student variants use a 21-frame KV window with 3 sink frames in both training and inference.

\paragraph{Evaluation.}
The 15s VBench evaluation uses Wan2.1-augmented prompts, with 5 videos generated per prompt.
The 60s VBench-Long evaluation uses all 1,003 entries in the original MovieGenBench prompt list without prompt expansion.
We generate 1 video per entry, with each rollout containing 1,005 frames.

\subsection{Out-of-sight World Modeling}
\label{sec:appendix-world-model}

\paragraph{Training.}
Training camera trajectories use left/right rotations or translations followed by a return to the initial viewpoint.
The learning rate is $1\times10^{-5}$, with LoRA rank 64, scaling parameter $\alpha=128$, and 0 dropout.
No gradient accumulation is used, giving an effective batch size of 8.
The EMA teacher is initialized from the student and updated after each optimization step with a decay of 0.9999.
For each training sample, we generate the student and fixed-view teacher rollouts online with shared initial noise and SDE transition noise.
The fixed-view teacher rollout uses a 12-frame KV window with 3 sink frames.
Since the world model performs image-to-video generation, both inputs used for historical supervision retain 1 sink frame corresponding to the initial image.
To reserve capacity for additional history within the model's native attention window of 12 latent frames, the shared recent context is reduced to 4 student-generated latent frames.
The history-conditioned input additionally includes up to 4 earlier latent frames from the fixed-view teacher rollout.

\paragraph{Inference.}
Evaluation uses the fine-tuned student weights.
The student uses a 12-frame KV window with 3 sink frames at $1280\times704$ resolution, generating 1 video per input.

\paragraph{Evaluation.}
All models are evaluated using the same initial images, prompts, preset camera trajectories, and random seeds.
Generation prompts describe the initial scene and its ongoing process, while task objectives are used only for evaluation.
For the PlayWorld Out-of-sight subset~\citep{ding2026playworld}, VLM-based scoring uses Doubao Seed 2.0 Lite.
Raw Mean averages ratings before instruction gating, and Gated Mean assigns a score of 1 to videos that fail the instruction check.
Evolution Quality measures the extent and causal plausibility of state evolution.
Subject Identity measures subject and appearance consistency across revisits.
Both metrics report mean 1--5 ratings before instruction gating.
For the MemoBench Real subset~\citep{chen2026memobench}, automated metrics are computed using the official MemoBench implementation.

\section{Additional Qualitative Analysis}
\label{sec:appendix-qualitative-analysis}
\label[appendix]{app:qualitative-analysis}

\begin{figure*}[p]
    \centering
    \includegraphics[width=\textwidth]{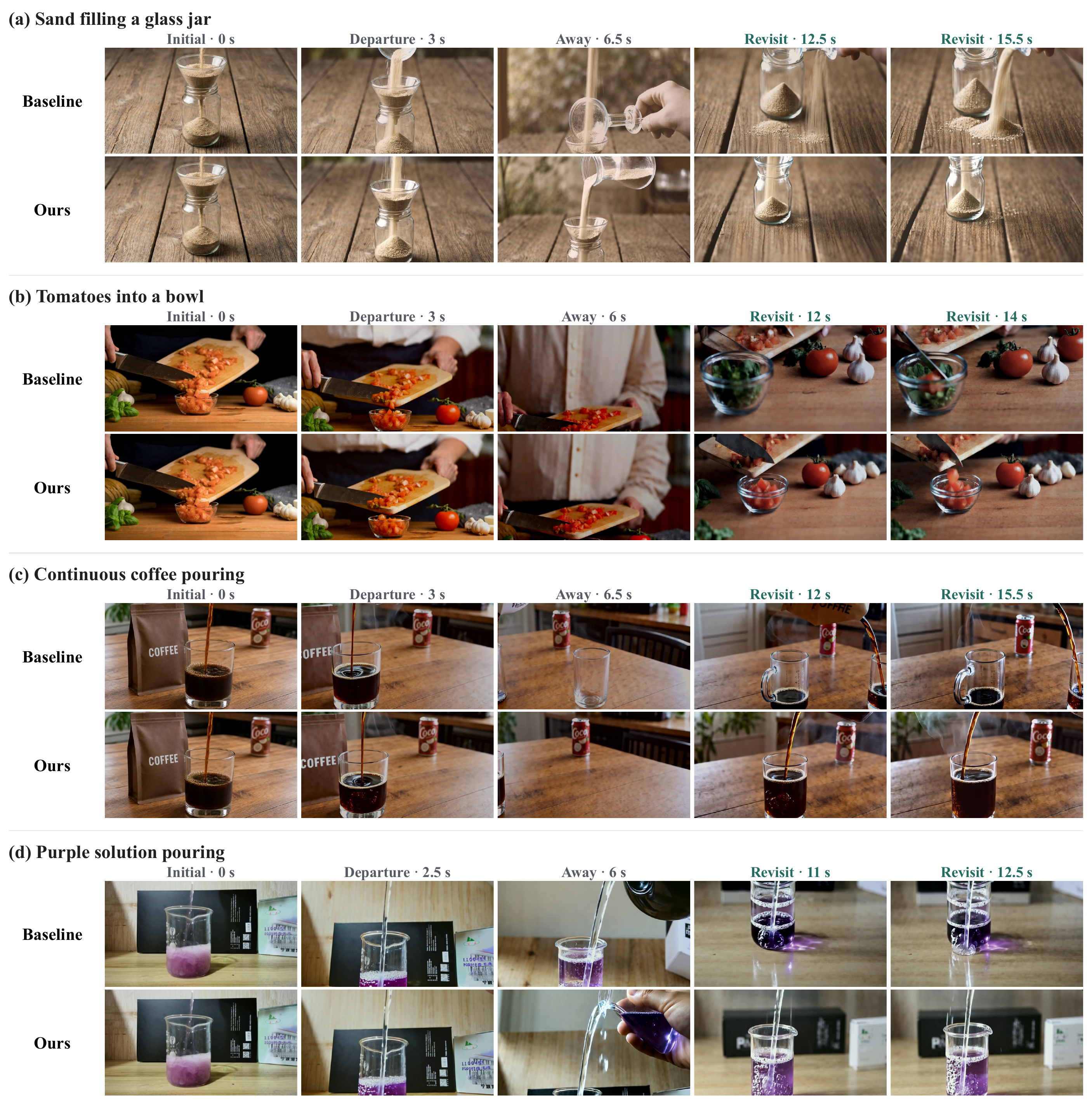}
    \caption{Additional qualitative comparisons of out-of-sight world modeling with the baseline and \method: \textbf{(a)} sand filling, \textbf{(b)} tomatoes into a bowl, \textbf{(c)} coffee pouring, and \textbf{(d)} purple solution pouring.}
    \label{fig:qualitative-comparison-appendix}
\end{figure*}

\noindent\textbf{Out-of-sight world modeling.}
\cref{fig:qualitative-comparison-appendix} shows improved object preservation and scene evolution upon revisit.
In the sand-filling example, the baseline accumulates sand outside the jar, while \method keeps the sand predominantly inside it.
\method retains the tomatoes in the bowl, preserves the drinking glass as the coffee level rises, and keeps the purple solution inside the beaker.
Some appearance and scene drift nevertheless remains.
The fixed-camera teacher rollouts used for historical supervision can themselves contain temporal errors and exhibit limited scene evolution, as discussed in \cref{sec:conclusion}.
These limitations reduce the accuracy of historical supervision and can contribute to inconsistencies upon revisit.

\begin{figure*}[p]
    \centering
    \includegraphics[width=\textwidth,height=0.9\textheight,keepaspectratio]{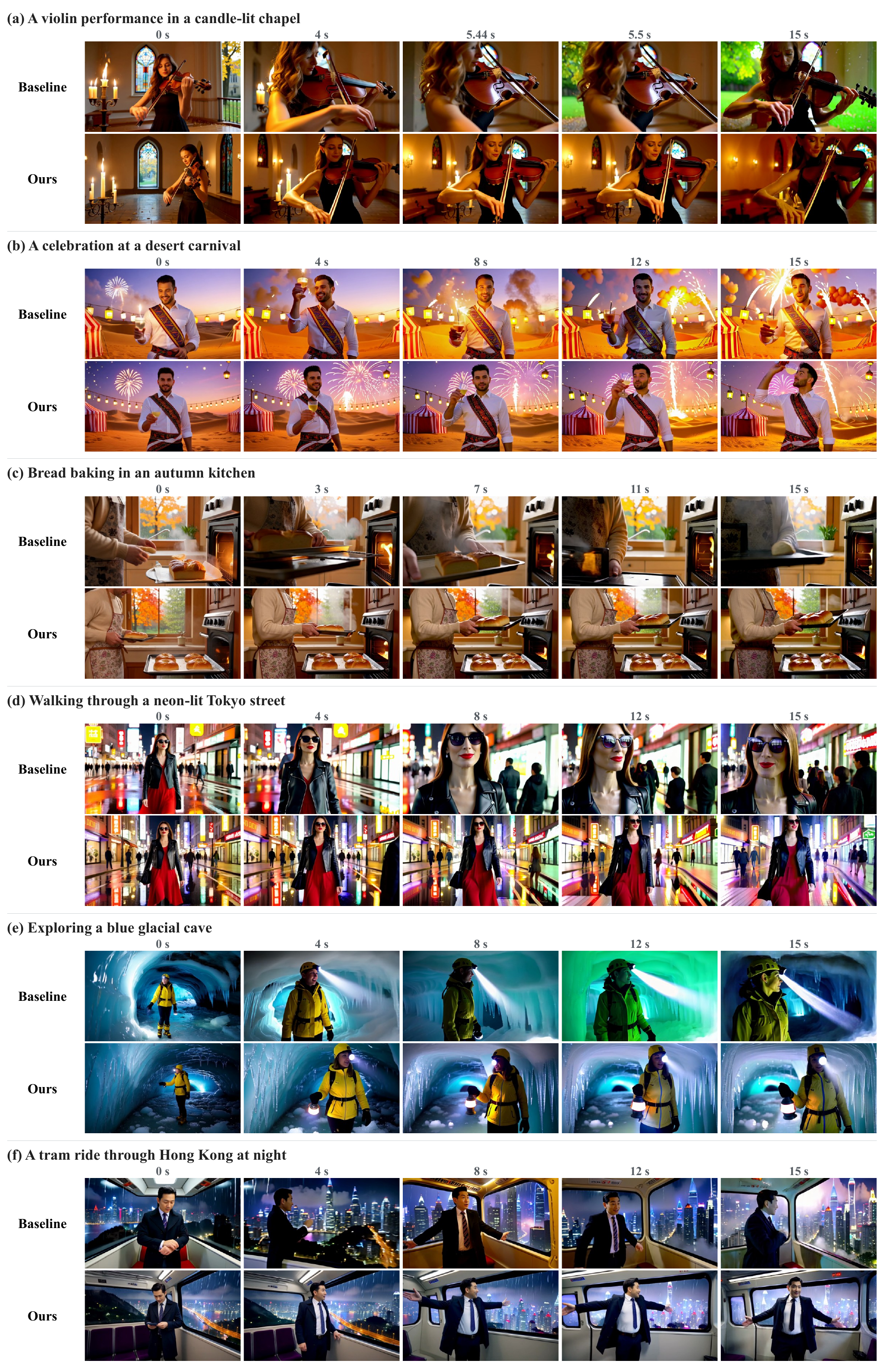}
    \caption{Additional qualitative comparisons of 15s video generation: \textbf{(a)} a violin performance in a candle-lit chapel, \textbf{(b)} a desert carnival, \textbf{(c)} bread baking, \textbf{(d)} a Tokyo street, \textbf{(e)} a glacial cave, and \textbf{(f)} a Hong Kong tram.}
    \label{fig:qualitative-comparison-15s}
\end{figure*}

\noindent\textbf{15-second video generation.}
\cref{fig:qualitative-comparison-15s} compares the DMD baseline and \method on six 15s examples: a violin performance in a candle-lit chapel, a desert carnival, bread baking, a walk through a Tokyo street, a glacial cave, and a Hong Kong tram.
In the chapel example (a), the baseline background changes abruptly from the chapel interior to outdoor greenery between the displayed frames at 5.44s and 5.5s.
In the carnival example (b), the baseline also changes the subject's appearance across frames.
Across these examples, \method better preserves the subjects, clothing, and surrounding scene at the sampled timestamps.

\begin{figure*}[p]
    \centering
    \includegraphics[width=\textwidth]{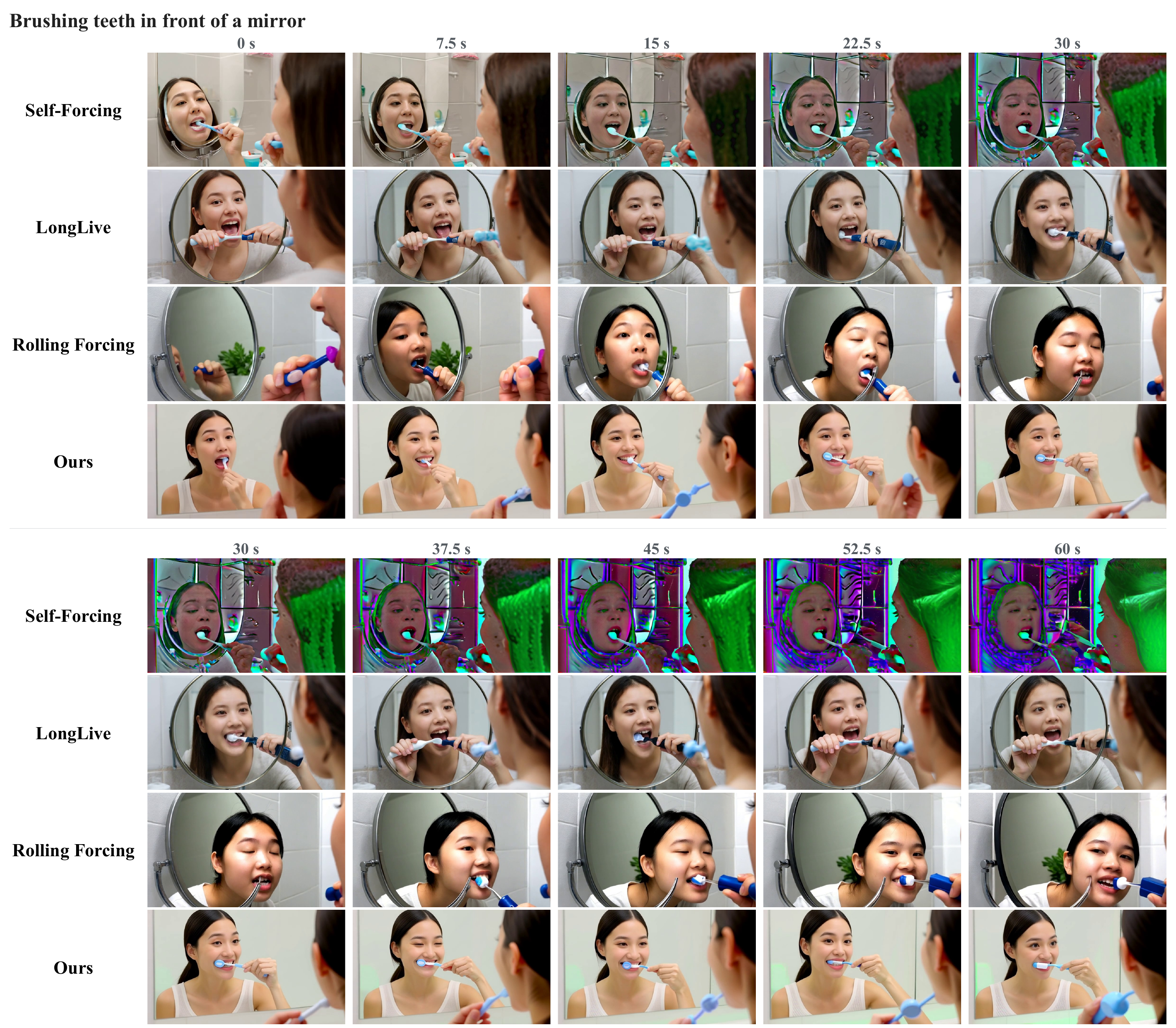}
    \caption{Additional qualitative comparisons of 60s video generation: brushing teeth in front of a mirror.}
    \label{fig:qualitative-comparison-60s}
\end{figure*}

\begin{figure*}[p]
    \centering
    \includegraphics[width=\textwidth]{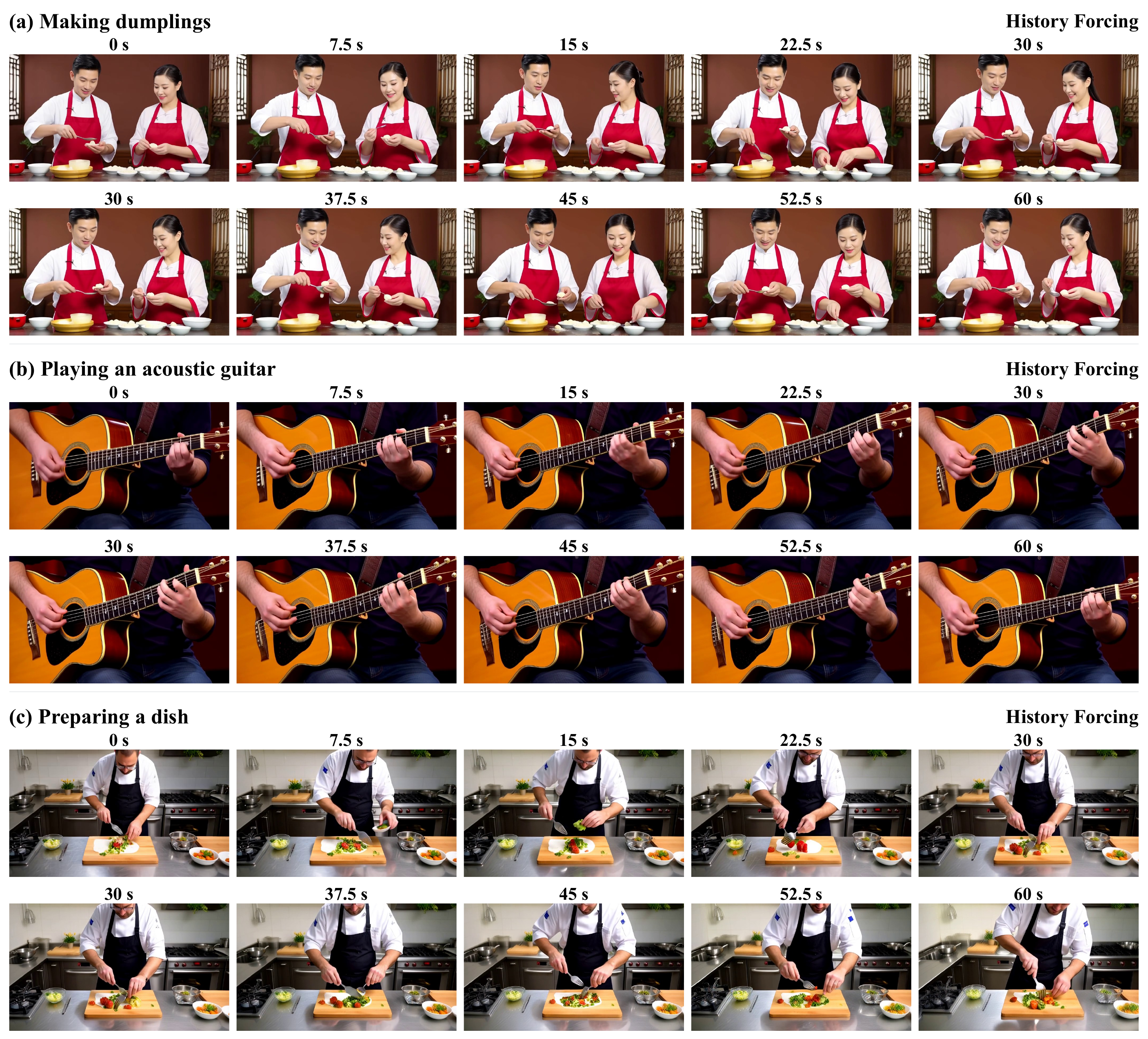}
    \caption{Additional qualitative examples of 60s video generation with \method: \textbf{(a)} making dumplings, \textbf{(b)} playing an acoustic guitar, and \textbf{(c)} a chef preparing a dish in a kitchen.}
    \label{fig:qualitative-examples-60s}
\end{figure*}

\noindent\textbf{60-second video generation.}
\cref{fig:qualitative-comparison-60s} compares \method with Self-Forcing, LongLive, and Rolling Forcing on a 60s toothbrushing example.
Self-Forcing develops severe color saturation and high-contrast artifacts as the rollout progresses.
LongLive produces inconsistent toothbrush geometry, with the brush appearing as an elongated object grasped at both ends in several frames.
Rolling Forcing distorts the mirror geometry: the reflected face extends beyond the mirror boundary, and the rim intersects the mouth and cheek.
\method maintains natural colors and more consistent toothbrush appearance and mirror geometry through 60s.
\cref{fig:qualitative-examples-60s} further shows that \method maintains consistent clothing, object appearance, and scene layout across the 60s rollouts.

\clearpage

\clearpage

\end{document}